\documentclass[letterpaper, 10 pt, conference]{ieeeconf}
\IEEEoverridecommandlockouts                              

\usepackage{times}
\usepackage{epsfig}
\usepackage{graphicx}
\usepackage{amsmath}
\usepackage{amssymb}
\usepackage{graphicx}
\usepackage{amsmath}
\usepackage{booktabs}
\usepackage{algorithm}
\usepackage{algorithmic}
\usepackage{comment}
\usepackage{graphicx}
\usepackage{amsmath,amssymb} 
\usepackage{color}
\usepackage{booktabs}
\usepackage{bm}

\usepackage[pagebackref=true,breaklinks=true,letterpaper=true,colorlinks,bookmarks=false]{hyperref}

\author{Jiakai Lin, Jinchang Zhang, and Guoyu Lu
\thanks{Jiakai Lin, Jinchang Zhang, Guoyu Lu are with the University of Georgia.
        {\tt\small guoyulu62@gmail.com}.}%
}

\begin{document}

\title{Keypoint Detection and Description for Raw Bayer Images}

\maketitle

\begin{abstract}
Keypoint detection and local feature description are fundamental tasks in robotic perception, critical for applications such as SLAM, robot localization, feature matching, pose estimation, and 3D mapping. While existing methods predominantly operate on RGB images, we propose a novel network that directly processes raw images, bypassing the need for the Image Signal Processor (ISP). This approach significantly reduces hardware requirements and memory consumption, which is crucial for robotic vision systems. Our method introduces two custom-designed convolutional kernels capable of performing convolutions directly on raw images, preserving inter-channel information without converting to RGB. Experimental results show that our network outperforms existing algorithms on raw images, achieving higher accuracy and stability under large rotations and scale variations. This work represents the first attempt to develop a keypoint detection and feature description network specifically for raw images, offering a more efficient solution for resource-constrained environments.
\end{abstract}

\vspace{-1mm}
\section{INTRODUCTION}
\label{sec:intro}
\vspace{-1mm}
Keypoint detection and local feature description play a critical role in the fields of robotics and computer vision, serving as fundamental components for a variety of tasks, such as Simultaneous Localization and Mapping (SLAM) \cite{mur2015orb}, localization \cite{lu2015localize}, motion estimation \cite{lu2023deep}, and 3D reconstruction \cite{lu2013large}. These tasks are essential for enabling robots to perceive, navigate, and interact with environments. Particularly, the frontend processing of keypoints is indispensable for accurate and efficient downstream applications, where the precision and robustness of keypoint detection and feature extraction directly impact the overall system performance.

Raw image data, directly output from the camera sensor, offers significant advantages over traditional RGB images in robotics tasks. By performing keypoint detection and feature description directly on raw images, we can bypass the Image Signal Processor (ISP), reducing hardware requirements and computational overhead. Additionally, raw images use only one-third of the memory compared to RGB images, making them highly beneficial for real-time applications on resource-constrained robots where efficient processing and minimal hardware dependency are crucial.
The development of keypoint detection and feature description has seen the emergence of classic algorithms like SIFT and ORB, as well as deep learning methods such as SuperPoint, DISK, and ALIKED. While these algorithms have been successful on RGB and grayscale images, their performance often degrades with significant changes in rotation or scale. Additionally, when processing raw images, current methods either convert them to RGB, introducing errors and requiring an ISP, or split them into four channels for convolution, neglecting important information exchange between channels.


We propose a novel network that performs keypoint detection and feature description extraction directly on raw images. By developing specialized convolutional kernels to process the raw image format, it preserves inter-channel information and avoids errors introduced by ISP processing. Operating directly on Bayer images not only retains the most original image information but also ensures the robustness of feature extraction, particularly in cases of rotation and scale variations, while also improving computational efficiency and resource utilization.
Existing methods that simply split Bayer images into four independent channels for convolution overlook the potential interaction between these channels. The distribution of each channel in a Bayer image is not independent but interrelated. As a result, channel separation leads to information loss, failing to capture the important relationships between the channels. Our proposed method, by designing specialized convolution kernels, operates directly on the Bayer image format, preserving and utilizing the interactions between channels, thereby enhancing the accuracy of feature extraction. The convolution kernels we introduce for Bayer images are specifically designed to handle the unique channel structure directly, without relying on RGB conversion or channel separation, avoiding the drawbacks of both. This enables more efficient and precise capture of geometric features in the image. Therefore, keypoint detection and descriptor extraction based on Bayer images is a reasonable and effective research direction, offering a better solution for robotics tasks on resource-constrained platforms.

The primary contributions of this paper are as follows:
1. We present the first method for keypoint detection and feature descriptor extraction tailored specifically for raw images.
2. We introduce novel convolutional kernels capable of directly handling the unique channel structure of raw images.
3. Our experimental results demonstrate superior performance compared to existing algorithms on raw images.
4. Our approach exhibits greater stability and accuracy when handling significant rotations and scale variations, outperforming current state-of-the-art methods. This work highlights the potential of raw image-based feature extraction, offering a more efficient and effective solution for robotic vision systems and paving the way for future advancements in real-time, resource-efficient visual processing.

\begin{figure*}[t]
\begin{center}
\includegraphics[width=16cm, height=7.6cm]{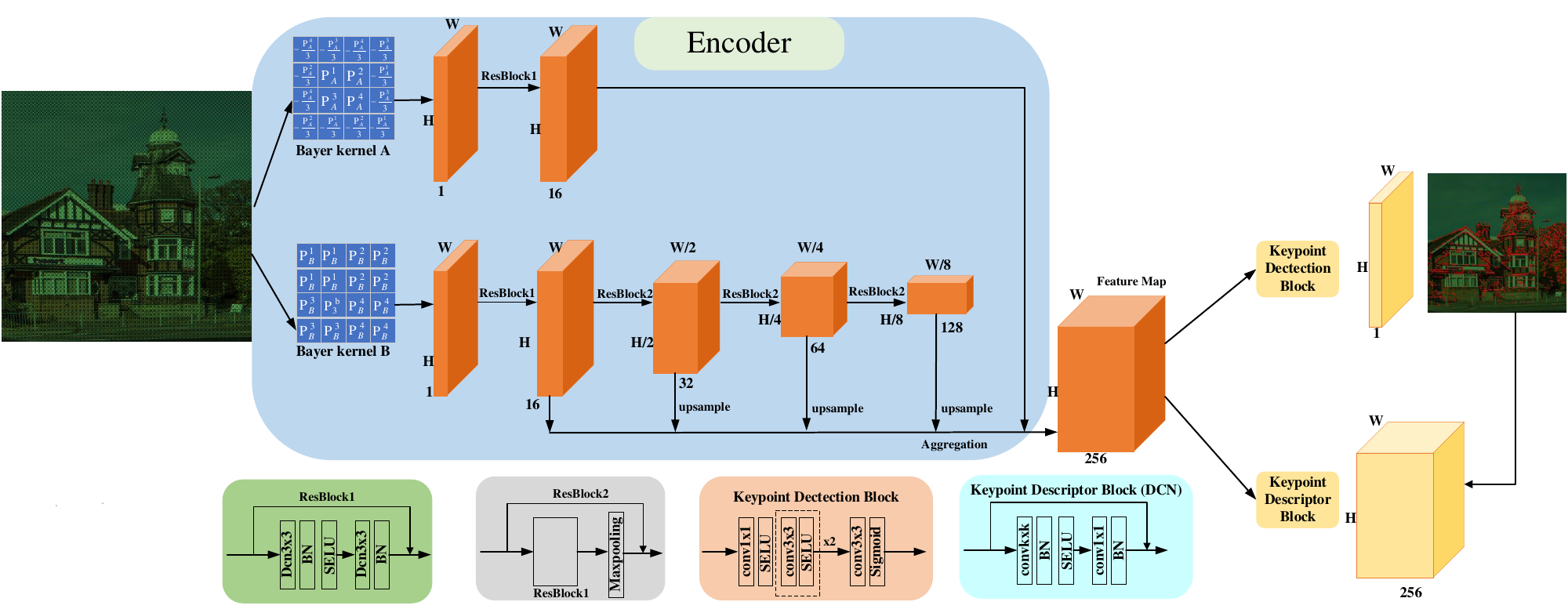}
\end{center}
\vspace{-7mm}
\caption{\textbf{Network architecture.} A raw image is processed through an Encoder and a Feature Pyramid Aggregation to generate a 256-dimensional feature map. This feature map is then passed through the Detection Block and Descriptor Block to obtain the Score map and Descriptor map.}

\label{fig:arch}
\vspace{-6mm}
\end{figure*}

\vspace{-1mm}
\section{RELATED WORK}
\vspace{-1mm}
\subsection{Extraction of Keypoint and Descriptor }
\vspace{-1mm}
The extraction of geometrically invariant descriptors is crucial in robotics and is typically defined by scale and orientation invariance \cite{schmid2000evaluation,mikolajczyk2005performance}. For example, SIFT \cite{lowe2004distinctive} constructs descriptors by estimating the scale and orientation of each keypoint in the scale space and extracting image patches. Similarly, FAST and ORB \cite{rublee2011orb} features achieve orientation invariance by rotating image patches. In terms of learning methods, descriptor extraction can be categorized into patch-based approaches and joint keypoint and descriptor learning methods. Patch-based methods, such as \cite{mishchuk2017working,han2015matchnet,balntas2016learning}, typically rely on data augmentation to achieve scale and orientation invariance. In contrast, joint learning methods like LIFT \cite{yi2016lift}, AffNet \cite{mishkin2018repeatability}, and LF-Net \cite{ono2018lf} use neural networks to estimate keypoint orientation and scale and apply these transformations to extract invariant descriptors.
Some studies focus on jointly estimating the score map and descriptor map, detecting keypoints from the score map, and sampling descriptors from the descriptor map. SuperPoint \cite{detone2018superpoint} introduced a lightweight network that generates training data using homographic adaptation. R2D2 \cite{revaud2019r2d2} detects keypoints by computing repeatability and reliability maps and trains descriptors using average precision loss.  \cite{suwanwimolkul2021learning} introduced a low-level feature detector into R2D2 to improve keypoint accuracy. ALIKE \cite{zhao2022alike} proposed a lightweight architecture with a precise keypoint detection module, enabling its application in real-time visual measurement tasks. ASLFeat \cite{luo2020aslfeat} improves localization accuracy and descriptor quality by using multi-level features and deformable convolutions, while D2D \cite{tian2020d2d} combines absolute and relative saliency for keypoint detection. ALIKED, leveraging the deformable nature of DCN \cite{zhu2019deformable}, introduces the SDDH module to efficiently extract geometrically invariant descriptors.Traditional keypoint and descriptor extraction methods, which rely on complete RGB images, perform poorly when applied to Bayer images. The main issues stem from the sparse pixel distribution, incomplete color information, and noise introduced during the demosaicing process. To address these challenges, we propose the Bayer convolution module, which operates directly on Bayer images, avoiding the information loss caused by demosaicing. This module captures local geometric and color features more effectively, enhancing the accuracy of keypoint detection and descriptor extraction, and successfully overcoming the limitations of traditional methods when processing Bayer images.

\vspace{-1mm}
\subsection{Raw Bayer Image}
\vspace{-1mm}
The Bayer pattern, typically using a Bayer Color Filter Array (CFA), is designed to support reliable demosaicing, with the primary goal of reconstructing a complete three-channel RGB color image by interpolating the missing red, green, and blue pixel values in the raw Bayer image \cite{li2008image,zhang2016universal,menon2011color,liu2020new}. To achieve this, researchers have proposed various handcrafted interpolation algorithms, including color-difference-based interpolation \cite{chung2006color}, edge-directed interpolation \cite{zhu2019deformable}, and reconstruction-based methods \cite{shao2014image}. With advances in technology, deep learning approaches have also been widely applied to improve demosaicing quality \cite{liu2020joint,tan2018deepdemosaicking,tan2017color,zhou2021image}. For instance, \cite{liu2020joint} introduced a self-guided network that uses an initially estimated green channel as guidance, employing edge loss to supervise network training, thereby restoring all missing pixel values in the raw image with reference to the ground truth color image. More high-level tasks, such as detection \cite{lu2023object} and segmentation \cite{Lu_2023_BMVC}, also have successfully utilized raw Bayer images. 

Nonetheless, the directly utilizing raw Bayer images for keypoint detection and descriptor extraction remains relatively unexplored in the field of robotics. The main challenge lies in the fact that raw Bayer images lack the complete color information available in traditional RGB images, making it more difficult to extract precise local features. Keypoint detection and descriptor extraction tasks rely on the distinctiveness and saliency of local regions, which require clear structural and texture information in the image. However, due to the pixel arrangement in Bayer images, many visual cues are not immediately apparent in their raw form, complicating the design of feature extraction algorithms tailored for Bayer images. These challenges limit the broader application of Bayer images in robotics tasks. This paper proposes a dedicated framework based on Bayer patterns, demonstrating how keypoint detection and descriptor extraction can be effectively performed on raw Bayer images and explores potential advantages in precise feature extraction.

\begin{figure}[t]
\begin{center}
\includegraphics[width=8.5cm, height=4.5cm]{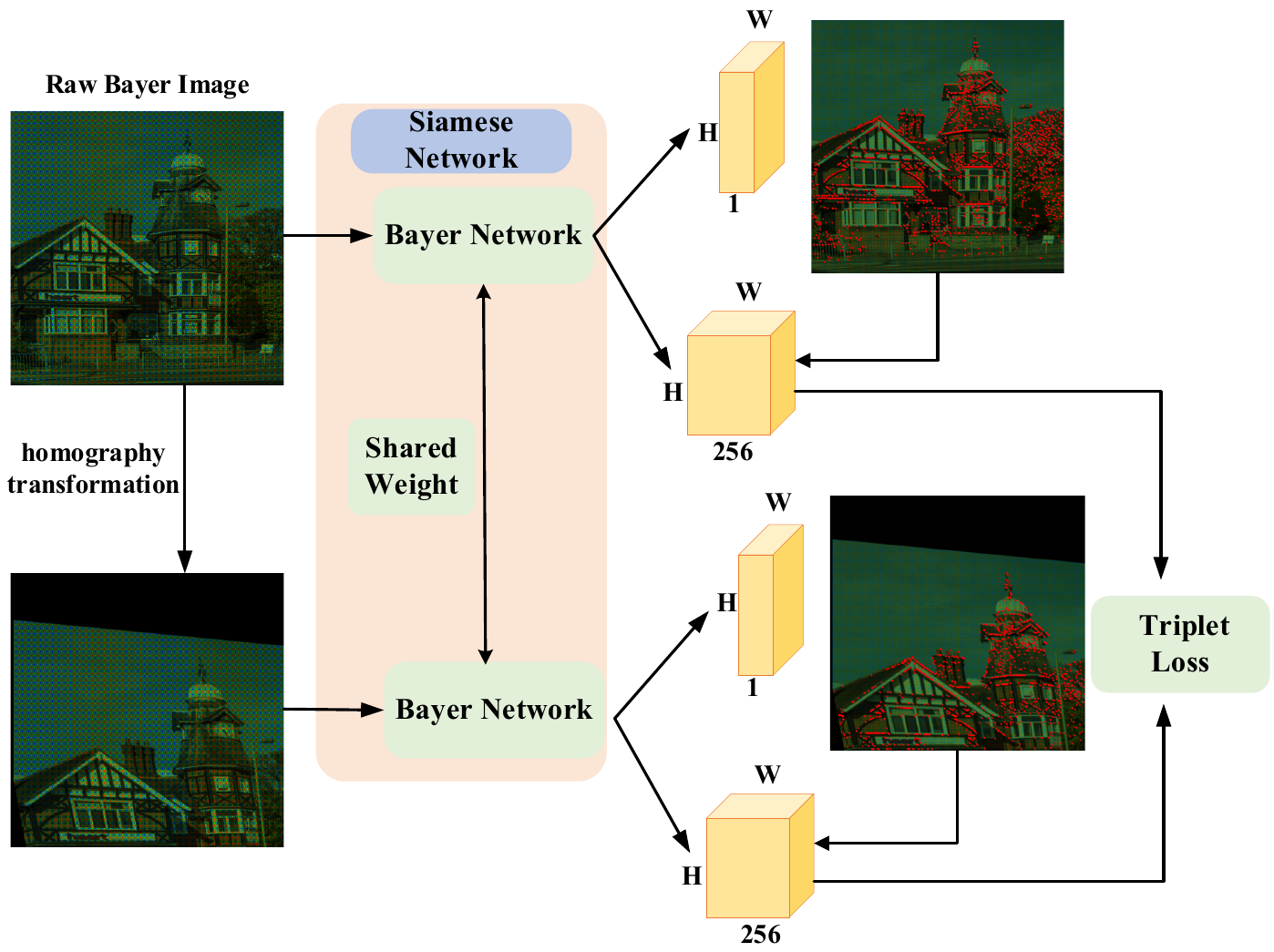}
\end{center}
\vspace{-6mm}
\caption{\textbf{Siamese Neural Network.} Input is a raw image pair with the known transformation. The network shares weights between the two branches, and weights are updated after each backpropagation.}
\label{fig:Siameseg}
\vspace{-3mm}
\end{figure}

\begin{figure}[t]
\begin{center}
\includegraphics[width=7cm, height=3cm]{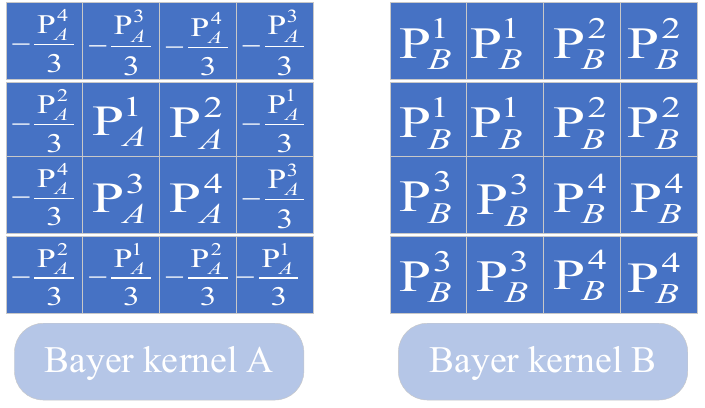}
\end{center}
\vspace{-6mm}
\caption{Two designed bayer convolution kernels.}
\label{fig:bayer kernel}
\vspace{-8mm}
\end{figure}

\vspace{-1mm}
\section{Keypoint Detection and Description}
\vspace{-1mm}
\subsection{Custom Bayer Convolution Kernels}
\vspace{-1mm}
Raw image data cannot be directly processed using traditional convolution due to the significant scale differences between the different channels (R, G, B). The sliding of the convolution kernel results in changes in the computed scale, as it is difficult to ensure that the number of pixels in each color channel remains consistent within the convolutional region. Additionally, during the upsampling or downsampling process, the convolution also struggles to maintain the structural consistency of the Bayer pattern. Therefore, directly applying convolutions to raw images often leads to information loss or unbalanced feature extraction. To address the challenges posed by raw image data, most existing approaches split the raw image according to the Bayer pattern into four channels (R, Gr, Gb, B), reducing the image from a $1 \times H \times W$ format to a $4 \times (H/2) \times (W/2)$ format. This separation allows convolution to be performed on each channel individually. However, this approach neglects the spatial correlations between neighboring pixels across different channels, which is particularly important for color representation in raw images. 

To solve these limitations, we derive formulas based on the Constant Color Difference Theory, which are adapted for convolution operations on raw images. Specifically, $R$ and $G$ represent the intensity of the corresponding pixel in the red and green channels. The theory can be expressed as:
\vspace{-3mm}
\begin{equation}
    R(i,j) - G(i,j) = R(i+\Delta i, j+\Delta j) - G(i+\Delta i, j+\Delta j)
    \vspace{-1mm}
\end{equation}
We transform this equation into a form suitable for convolution, applicable to any two channels in an RGB image:
\vspace{-2mm}
\begin{footnotesize}
\begin{equation}
\vspace{-1mm}
    p_a [R(i,j) - R(i+\Delta i,j+\Delta j)] = p_a [G(i,j) - G(i+\Delta i,j+\Delta j)]
    \label{eq2}
    \vspace{-6mm}
\end{equation}
\end{footnotesize}

\begin{footnotesize}
\begin{equation}
    p_b [R(i,j) + G(i+\Delta i,j+\Delta j)] = p_b [R(i+\Delta i,j+\Delta j) + G(i,j)]
    \label{eq3}
    \vspace{-2mm}
\end{equation}
\end{footnotesize}
Here, $p_a$ and $p_b$ are learnable parameters acting as convolution filters applied to any combination of two RGB channels, capturing both color differences and intensity variations.
Based on these equations, we design two specific $4 \times 4$ Bayer convolution kernels, as shown in Figure \ref{fig:bayer kernel}. These kernels contain 16 positions, of which only four are independent parameters ($p_1$, $p_2$, $p_3$, and $p_4$), while the remaining 12 positions are constrained by these parameters. Kernel 1, based on equation\eqref{eq2}, primarily captures local color variation, while Kernel 2, based on equation\eqref{eq3}, focuses on learning color intensity and related information.

After applying our designed convolution kernels, the resulting feature map for each pixel achieves consistent scaling across all color channels. This consistency enables the feature map to seamlessly integrate with traditional convolutional operations, including those designed for RGB images, such as standard convolutions, upsampling, and other network operations used in deep learning frameworks.

\vspace{-1mm}
\subsection{Feature Encoder}
\vspace{-1mm}
The input is a RAW Bayer image, and the network structure consists of two independent Bayer convolution layers. The first convolution layer is specifically designed to extract relevant information from the image, while the second convolution layer extracts standard convolution features from the Bayer image. The relevant information is processed through a residual block using $3\times3$ DCNs (Deformable Convolutional Networks) \cite{zhu2019deformable}, generating a $16 \times W  \times H$ feature map, \(C1\). Simultaneously, the features extracted from the standard convolution path of the Bayer image also pass through a residual block with $3\times3$ DCNs, producing a 16×W×H feature map, \(F1\).
To expand the receptive field and improve computational efficiency, the second module consists of a residual block made up of a $3\times3$ DCN followed by $2\times2$ max pooling, which downsamples \(F1\) to produce a $32\times \left( {W/2} \right) \times \left( {H/2} \right)$ feature map, \(F2\). The subsequent third and fourth modules follow the same structure, further downsampling to generate $64 \times \left( {W/4} \right) \times \left( {H/4} \right)$ feature map \(F3\) and $128 \times \left( {W/8} \right) \times \left( {H/8} \right)$ feature map \(F4\), respectively.
In the feature aggregation part, the network fuses multi-scale features \(\{C1, F1, F2, F3, F4\}\) to enhance localization and representation capabilities. By performing a series of upsampling and feature alignment operations, the aligned features \(\{C1, F1, Fu2, Fu3, Fu4\}\) are concatenated, resulting in a final aggregated feature map, \(F\), with a size of $256 \times W\times H $, which is used for keypoint detection and descriptor extraction.

\vspace{-1mm}
\subsection{Keypoint Detection and  Descriptor Block}
\vspace{-1mm}
\textbf{Keypoint Detection Block: }
We designed a keypoint detection block that leverages the aggregated feature \( F \) to estimate the score map. The Keypoint Decoder first reduces the feature channels to 8 through a 1 × 1 convolution layer, followed by two 3 × 3 convolution layers for further feature encoding. Finally, the score map \( S \) is generated using a 3 × 3 convolution layer and a Sigmoid activation function, ensuring that the keypoint detection results exhibit an appropriate probability distribution.

\textbf{Keypoint Descriptor Block}
For the descriptor block, we use Deformable Convolution (DCN) \cite{zhu2019deformable} to extract features from the aggregated feature map \( F \), generating a descriptor feature map of size \( H \times W \times 256 \). By introducing learnable spatial offsets in the deformable convolution to adjust the sampling locations of the convolution kernel, the descriptor decoder can better capture non-rigid geometric variations in the input image.
The deformable convolution includes an offset branch that learns a deformable offset \( \Delta p \). This offset determines the specific sampling locations of the convolution kernel at each position on the input feature map. Through the offset estimation network, we obtain an offset matrix of size \( K \times K \), where \( K \) is the size of the convolution kernel. In the deformable convolution operation, the convolution kernel dynamically adjusts the sampling points at each position based on the estimated offset \( \Delta p \).
For each pixel \( p \) on the feature map \( F \), the deformable convolution is computed as:
\vspace{-3mm}
\begin{equation}
y(p) = \sum_{k=1}^{K^2} w_k \cdot F(p + p_k + \Delta p_k)
\vspace{-3mm}
\end{equation}
\noindent where \( w_k \) is the convolution kernel weight, \( p_k \) is the standard sampling location of the convolution kernel, and \( \Delta p_k \) is the learned offset.
Through stacking multiple deformable convolutions and aggregating features, we transform the input feature map \( F \) into a feature map \( D \in \mathbb{R}^{H \times W \times 256} \), which is L2-normalized to obtain the descriptor map \( D \) of size \( H \times W \times \text{dim} \).

\begin{figure*}[t]
\begin{center}
\includegraphics[width=17.5cm, height=5cm]{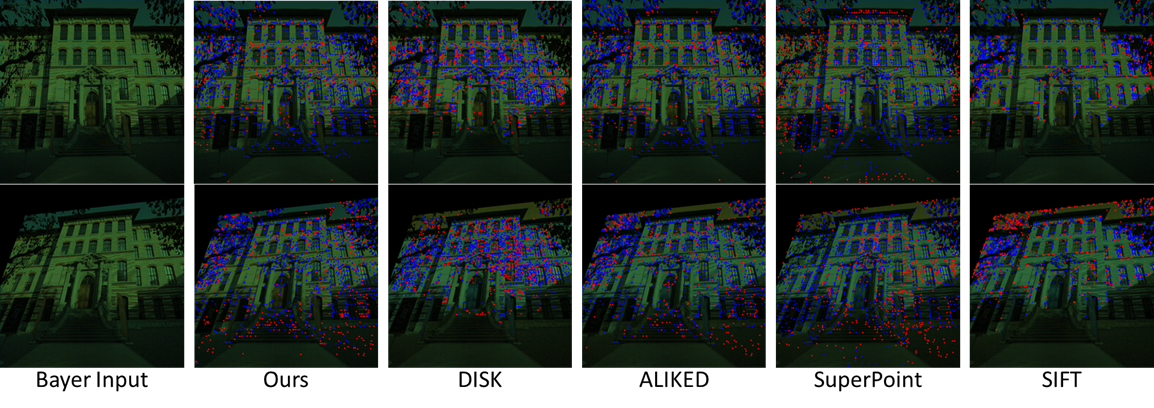}
\end{center}
\vspace{-6mm}
\caption{\textbf{Keypoints repeatability results.} From left to right: the input Bayer image pairs and the outputs of our model, DISK~\cite{tyszkiewicz2020disk}, ALIKED~\cite{zhao2023aliked}, SuperPoint~\cite{detone2018superpoint},  and SIFT~\cite{lowe2004distinctive}. Blue keypoints indicate those that meet the repeatability criterion ($\epsilon = 3$), while red keypoints represent those that do not.}
\label{fig:repeat}
\vspace{-6mm}
\end{figure*}

\vspace{-1mm}
\subsection{Keypoint Detection Training Process and Loss}
\vspace{-1mm}
\textbf{KeyPoint training}: We use the synthetic dataset provided by \cite{detone2018superpoint} - Synthetic Shapes, which contains simplified 2D geometric shapes such as quadrilaterals, triangles, line segments, and ellipses rendered through synthetic data. We pre-train our keypoint network on the Synthetic Shapes dataset while freezing the descriptor head during this phase.
First, we apply multiple random homographic transformations \( H_i \) to each image \( I \), generating several images from different perspectives and scales. This ensures that the model can learn features under various viewpoints and scales. Specifically, we sample translation, scaling, rotation, and symmetric perspective distortion from a truncated normal distribution within a predetermined range. These transformations are combined with center cropping to avoid artifacts at the image boundaries. For each image \( I_i \) after applying the random homography \( H_i \), our network generates a keypoint probability map (score map) \( S_i(x, y) \), where the value at each pixel indicates the likelihood of it being a keypoint.
Ideally, a good keypoint detector should exhibit covariance with homographies, meaning the detected keypoints should remain consistent under different homographic transformations. However, in practice, the detector does not fully achieve covariance — different homographies \( H_i \) lead to the detection of different keypoints \( x \). To enhance the robustness of the detector, we compute the empirical sum over random samples of \( H_i \) to obtain a more robust keypoint detection result.
For each original image \( I \), we apply the inverse transformation \( H_i^{-1} \) to the probability map \( S_i(x, y) \) generated from the homography-transformed image \( I_i \), mapping it back to the original image coordinate system. The formula is given by:
\vspace{-2mm}
\begin{equation}
\vspace{-2mm}
S'_i(x, y) = S_i(H_i^{-1}(x, y))
\end{equation}
\noindent where \( H_i^{-1}(x, y) \) maps the pixel \( (x, y) \) from the original image back to the corresponding location in the transformed image using the inverse homography.
For each original image, we apply \( N \) random homographic transformations \( H_i \), and then average the probability maps \( S'_i(x, y) \) of all inverse transformed images to generate the final pseudo-ground truth keypoint probability map \( S_{\text{pse}}(x, y) \):
\vspace{-2mm}
\begin{equation}
\vspace{-2mm}
S_{\text{pse}}(x, y) = \frac{1}{N} \sum_{i=1}^{N} S'_i(x, y) = \frac{1}{N} \sum_{i=1}^{N} S_i(H_i^{-1}(x, y))
\end{equation}
This averaged probability map can be regarded as the supervision signal for the pseudo-ground truth keypoints, where the average probability value at each pixel represents the probability distribution of whether it is a keypoint. This probability map is used as the ground truth during the network's training process, guiding the network for supervised learning. By applying multiple geometric transformations and aggregating the probabilities, this method mitigates the instability caused by noise or transformations in a single image, resulting in more robust pseudo-ground truth keypoints.

\textbf{keypoint detection:} In the keypoint detection network, we adopt the Binary Cross-Entropy Loss. Each pixel is treated as a binary classification problem, where the goal is to predict whether the pixel is a keypoint. This loss function measures the difference between the predicted keypoint probability map and the pseudo-label, guiding the network to learn effective keypoint detection. The network generates a predicted probability \( S_{\text{pred}}(h, w) \) for each pixel, which represents the probability of that pixel being a keypoint. The keypoint detection loss function can be defined as:
\begin{align}
\nonumber   
\mathcal{L}_{\text{k}} = - \frac{1}{H W} \sum_{h=1}^{H} \sum_{w=1}^{W} \Big[ {S_{\text{pse}}}(h, w) \log S_{\text{pred}}(h, w) \\
+ (1 - {S_{\text{pse}}}(h, w)) \log (1 - S_{\text{pred}}(h, w)) \Big]
\end{align}

 \( H \), \( W \) is the height and width of the image.
- \( S_{\text{pred}}(h, w) \) is the predicted probability that pixel \( (h, w) \) is a keypoint.

\textbf{Dissipation Peak Loss:}: It can better handle the spatial distribution problem in keypoint detection. In the score map of keypoints, ideally, the true keypoints should have a high score, while the surrounding pixels should have relatively lower scores. We need to consider the spatial distribution of the scores, i.e., the relationship between each pixel's position relative to the detected keypoint and its score.
The loss constrains the spatial distribution by calculating the distance between surrounding pixels and the keypoint. For an \( N \times N \) score block, we first compute the Euclidean distance \( d_{i,j} \) between each pixel \( (i, j) \) and the keypoint at \( (x_k, y_k) \):
$d_{i,j} = \sqrt{(i - x_k)^2 + (j - y_k)^2}$. 
Next, we define the relationship between the distance and the score to construct the loss function. The Dissipation Peak Loss is defined as:
\vspace{-2mm}
\begin{equation}
\vspace{-2mm}
  \mathcal{L}_{\text{peak}} = \sum_{i=1}^{N} \sum_{j=1}^{N} d_{i,j} \cdot S(i,j)
\end{equation}
\noindent where \( d_{i,j} \) is the distance between pixel \( (i, j) \) and the keypoint. \( S(i,j) \) is the score of pixel \( (i,j) \).
This loss function multiplies the distance by the score to ensure that the keypoint score is significantly higher than its surroundings, 
 controlling the spatial distribution of the score map.

\vspace{-1mm}
\subsection{Keypoint Description Training Process and Loss }
\vspace{-1mm}
We use a siamese network composed of two BayerNet models with shared parameters shown in Figure \ref{fig:Siameseg} to extract keypoints and descriptors from two images. The relationship between the images is defined by a known homography transformation, allowing us to map the location of the keypoints and descriptors. We employ the Triplet Loss to compare the descriptors, aiming to make the matching descriptors as similar as possible, while making non-matching descriptors as distinct as possible.
We apply a random homography transformation \( H \) to the original image \( I_1 \), generating the transformed image \( I_2 \). There is a well-defined geometric relationship between these two images.
A Siamese network composed of two BayerNet models with shared parameters is used to extract keypoints and descriptors from \( I_1 \) and \( I_2 \) independently
The descriptors are denoted as: \( \mathbf{D}_1^i \): The descriptor of the \( i \)-th keypoint in image \( I_1 \) and \( \mathbf{D}_2^j \): The descriptor of the \( j \)-th keypoint in image \( I_2 \).
Based on the known homography transformation \( H \), the \( i \)-th keypoint in \( I_1 \), \( \mathbf{p}_1^i \), is mapped to the corresponding point \( \mathbf{p}_2^i = H(\mathbf{p}_1^i) \) in \( I_2 \). Then, we calculate the Euclidean distance (L2 distance) between the descriptors of these matching keypoints:
\vspace{-1mm}
\begin{equation}
   d_{\text{positive}} = \left\| \mathbf{D}_1^i - \mathbf{D}_2^{H(i)} \right\|_2 
\vspace{-1mm}
\end{equation}
This distance is minimized as it represents the similarity between the descriptors of the same keypoint in both images.
For each keypoint \( \mathbf{p}_1^i \) and its corresponding descriptor \( \mathbf{D}_1^i \), we randomly select the descriptor \( \mathbf{D}_2^j \) at the position \( \mathbf{P}_2^j = H(\mathbf{P}_1^j) \) in image \( I_2 \), corresponding to another keypoint from image \( I_1 \). We then compute the Euclidean distance for this non-matching descriptor:
\vspace{-1mm}
\begin{equation}
d_{\text{negative}} = \left\| \mathbf{D}_1^i - \mathbf{D}_2^j \right\|_2
\vspace{-1mm}
\end{equation}
This distance is maximized as it represents the dissimilarity between descriptors of non-matching keypoints. The goal of the descriptor loss is to minimize the distance between positive samples \( d_{\text{positive}} \) while maximizing the distance between negative samples \( d_{\text{negative}} \). A margin \( \alpha \) is introduced to enforce a separation between positive and negative pairs. The triplet loss is defined as:
\vspace{-2mm}
\begin{equation}
    \mathcal{L}_{\text{triplet}} = \sum_i \left[ \left\| \mathbf{D}_1^i - \mathbf{D}_2^{H(i)} \right\|_2^2 - \left\| \mathbf{D}_1^i - \mathbf{D}_2^k \right\|_2^2 + \alpha \right]_+
    \vspace{-1mm}
\end{equation}
\noindent where \( [ \cdot ]_+ \) denotes the positive part, meaning the result is zero if the value inside the bracket is negative.
\( \mathbf{D}_1^i \) and \( \mathbf{D}_2^{H(i)} \) represent the positive sample pair, meaning they are descriptors of the same keypoint in both images. \( \mathbf{D}_1^i \) and \( \mathbf{D}_2^k \) represent the negative sample pair, meaning they are descriptors of different keypoints.
\( \alpha \) is the margin that separates positive and negative pairs.
The triplet loss helps the network learn to generate discriminative descriptors, bringing descriptors of the same keypoint closer together while pushing descriptors of different keypoints further apart.

\begin{table}[t]
\centering
\resizebox{0.45\textwidth}{!}{
\begin{tabular}{l|ccccc}
\hline
 & \textbf{Our Model} & \textbf{ALIKED} & \textbf{DISK} & \textbf{SuperPoint} & \textbf{SIFT} \\ \hline
\textbf{Illumination (RAW) }  
& \textcolor{red}{68.02\%} & \textcolor{blue}{66.70\%} & 65.73\% & 58.41\% & 49.78\% \\ \hline
\textbf{Illumination (RGB) }  
& -
& 70.23\% 
& 69.12\% 
& 63.30\% 
& 54.02\% \\ \hline
\textbf{Viewpoint (RAW)}   
 & \textcolor{red}{54.31\%} & \textcolor{blue}{52.29\%} & 50.15\% & 43.64\% & 40.70\% \\ \hline
\textbf{Viewpoint (RGB)}   
& -
& 56.01\% 
& 54.79\% 
& 49.65\% 
& 44.71\%  \\ \hline
\end{tabular}
}
\vspace{-1.5mm}
\caption{\textbf{HPatches Detector Repeatability($\epsilon = 3$).} Results in 57 illumination scenes and 59 viewpoint scenes. The best result in each row of raw image is highlighted in red, and the second best in blue.}
\label{Repeatability}
\vspace{-7mm}
\end{table}

\begin{table}[t]
\centering
\begin{tabular}{l|ccc}
\hline
\textbf{Model} & \textbf{MMA@5} & \textbf{MHA@5} & \textbf{MS@5} \\ 
\hline
SIFT        & 54.87\% & 60.01\% & 32.77\% \\
SuperPoint  & 60.34\% & 65.91\% & 46.77\% \\
DISK        & \textcolor{red}{72.46\%} & 66.12\% & \textcolor{blue}{55.63\%} \\
ALIKED      & 69.33\% & \textcolor{blue}{72.50\%} & 55.49\% \\
\hline
\textbf{Our Model}   & \textbf{\textcolor{blue}{70.80\%}} & \textbf{\textcolor{red}{74.17\%}} & \textbf{\textcolor{red}{57.11\%}} \\
\hline
\end{tabular}
\vspace{-1.5mm}
\caption{\textbf{HPatches Homography Estimation}. Results for MMA, MHA, and MS on raw images. The best result for each metric is highlighted in red, and the second-best result is in blue.}
\label{tab:metrics}
\vspace{-10mm}
\end{table}

\begin{figure*}[t]
\begin{center}
\includegraphics[width=17.5cm, height=6cm]{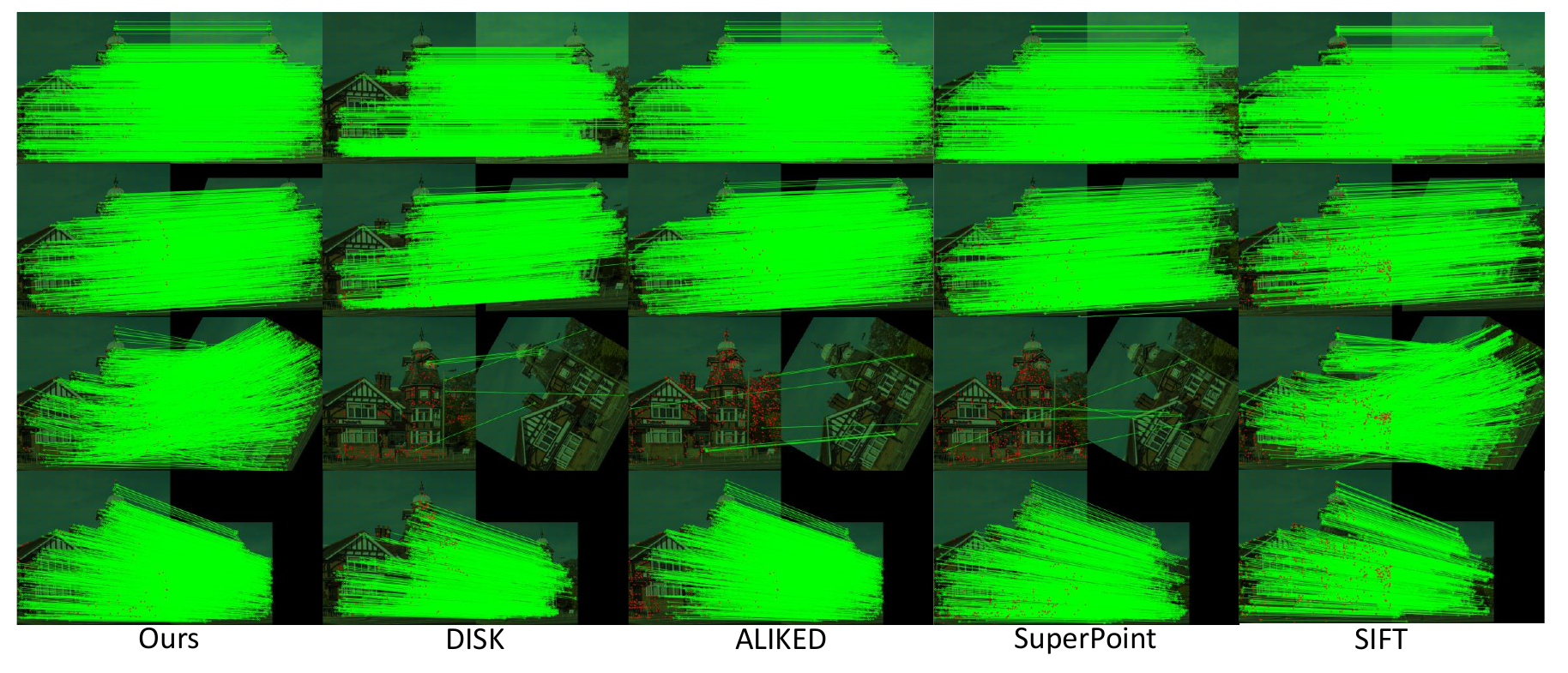}
\end{center}
\vspace{-7mm}
\caption{\textbf{Qualitative Results of Deformable Invariance Evaluation.} Green lines represent correct correspondences under RANSAC ($\epsilon = 5$), while red points indicate incorrect matching after filtering. Row 1 corresponds to exposure changes, Row 2 shows H-perspective transformations, Row 3 demonstrates large-angle rotations, and Row 4 illustrates scale changes. From left to right, the columns represent the bayer input and the results from our model, DISK~\cite{tyszkiewicz2020disk}, ALIKED~\cite{zhao2023aliked}, SuperPoint~\cite{detone2018superpoint}, and SIFT~\cite{lowe2004distinctive}.}
\label{fig:compare}
\vspace{-7mm}
\end{figure*}
\vspace{-1mm}
\section{Experiments}
In this section we compare quantitative results of the methods presented in the paper. The evaluation metric for the keypoint detection task is repeatability, while the evaluation tasks for descriptor extraction are homography estimation and deformable invariance.
\vspace{-2mm}
\subsection{HPatches Repeatability}
The evaluation is performed on the HPatches dataset \cite{balntas2017hpatches}. It consists of 116 scenes, where the first 57 scenes exhibit large changes in illumination and the remaining 59 scenes feature significant viewpoint changes. Each scene contains 6 images, providing a diverse testing set for evaluating the repeatability of keypoint detection and homography estimation of descriptor extraction. For our experiments, we extracted the raw image input from RGB images  following the Bayer pattern.
We tested keypoint repeatability on our model, several state-of-the-art (SOTA) models (ALIKED~\cite{zhao2023aliked}, DISK~\cite{tyszkiewicz2020disk}, SuperPoint~\cite{detone2018superpoint}), and the classical SIFT~\cite{lowe2004distinctive}, all with a maximum keypoint limit set to 2048 and thereshold $\epsilon = 3$ . 

The experimental results shown in table \ref{Repeatability} and figure \ref{fig:repeat} illustrate that our model outperforms existing SOTA models when raw images are used as input, achieving the highest repeatability. Moreover, the gap between our model and the SOTA models using RGB as input is only about 2\%, even though we only utilize one-third of the channel information. 
Additionally, we generate the simulated Bayer pattern raw image by directly cropping the channels from existing RGB images. This means that the channel information we use has already been processed by the ISP (Image Signal Processor) and is not truly raw. If we were to use genuine raw images without ISP processing, the performance gap would likely be even smaller. This experiment demonstrates that our model performs exceptionally well in keypoint detection tasks on raw images, highlighting its capability to deliver strong results even with reduced input data dimensions.

\vspace{-2mm}
\subsection{Homography Estimation}
\vspace{-1mm}
We evaluated the homography estimation performance of models on the 116 scenes from the HPatches dataset, where each scene contains 5 images and their respective homography matrices relative to a reference image as ground truth. The evaluation was conducted using three metrics: Mean Matching Accuracy (MMA), Mean Homography Accuracy (MHA), and Matching Score (MS). For all models, up to 2048 keypoints are extracted with a threshold of 0.1 on the score map and descriptors are matched using BFMatcher. The homography matrix was computed using RANSAC with a threshold $\epsilon = 5$.
The results in table \ref{tab:metrics} show that our model achieves the highest MHA and MS on raw images. Although our model's MMA is slightly behind DISK, DISK's MHA is significantly lower than that of our model and ALIKED. In homography estimation tasks, MHA is more important than MMA. In the MScore evaluation, our model performed the best with a slight advantage alongside ALIKED and DISK.

\vspace{-2mm}
\subsection{Deformable Invariance Evaluation}
\vspace{-1mm}
In this evaluation, we applied four different types of random transformations to the test images to simulate various real-world conditions: exposure changes (1.3 to 2 times intensity), homography adaptation (translation and perspective changes within ±30\%), large-angle rotations (45-90 degrees), and scale transformations (0.6 to 1.4 times). For each test, the maximum number of keypoints was set to 2048. The descriptors extracted from the keypoints were matched using OpenCV's BFMatcher with L2 distance, and the matches were evaluated using `findHomography` with a threshold of 5 pixels. Figure \ref{fig:compare} shows an example of the test results on one raw image. These metrics show that our model achieved the best performance on the homography estimation task for raw images.
The results demonstrate that our model exhibits strong stability, particularly in the case of large-angle rotations on raw images, significantly outperforming all other models in rotation invariance. Additionally, throughout the transformations, the proportion of inliers remained the highest for our model, indicating superior robustness.

\vspace{-1mm}
\section{Conclusion}
\vspace{-1mm}
In this work, we present the first method for keypoint detection and descriptor extraction specifically designed for raw Bayer images. Our approach introduces novel convolution kernels that operate directly on raw image data, eliminating the need for pre-processing into the RGB format. Furthermore, we propose the first 256-dimensional pixel-wise descriptor tailored for raw images. Our method surpasses state-of-the-art algorithms across various raw image matching tasks, delivering substantial improvements, especially in challenging cases involving image pairs with significant rotational differences. This advancement paves the way for future applications of raw image processing in diverse downstream tasks, providing a robust solution for raw image-based pipelines.

\textbf{Acknowledgment}: This work is supported by NSF Grants NO. 2340882, 2334624, 2334246, and 2334690. 

{
\bibliographystyle{ieee_fullname}
\bibliography{egbib}
}
\end{document}